\documentclass[twoside]{article}

\usepackage[accepted]{aistats2019}
\usepackage[numbers, sort&compress]{natbib}
\usepackage[utf8]{inputenc} 
\usepackage[T1]{fontenc}    
\usepackage{hyperref}       
\usepackage{url}            
\usepackage{booktabs}       
\usepackage{amsfonts}       
\usepackage{nicefrac}       
\usepackage{microtype}      

\usepackage{url}

\usepackage{breakurl}
\usepackage[font=small]{caption}
\usepackage{wrapfig}

\usepackage{multirow}
\usepackage{dsfont}
\usepackage{multicol}
\usepackage{amsmath}
\usepackage{amsthm}
\usepackage{amssymb}
\usepackage{bm}
\usepackage{mathrsfs}
\usepackage{xcolor}
\usepackage{colortbl}
\usepackage{subcaption}
\usepackage{mathtools}
\usepackage{nccmath}
\usepackage[hang,flushmargin]{footmisc}

\usepackage{color}
\usepackage{soul}

\usepackage{pgfplots}
\pgfplotsset{
  compat=newest,
  plot coordinates/math parser=false,
  tick label style={font=\footnotesize, /pgf/number format/fixed},
  label style={font=\small},
  legend style={font=\small},
  every axis/.append style={
    tick align=outside,
    clip mode=individual,
    scaled ticks=false,
    thick,
    tick style={semithick, black}
  }
}

\pgfkeys{/pgf/number format/.cd, set thousands separator={\,}}

\usetikzlibrary{calc}
\usepgfplotslibrary{external}
\usetikzlibrary{external}
\newlength\smallfigurewidth
\newlength\smallfigureheight
\newlength\onecolumnsquarefigurewidth
\newlength\onecolumnsquarefigureheight
\newlength\figurewidth
\newlength\figureheight
\newcommand{\mc}[1]{\mathcal{#1}}
\newcommand{\data}{\mc{D}}

\newcommand{\given}{\mid}
\newcommand{\E}{\mathbb{E}}
\newcommand{\intd}[1]{\,\mathrm{d}{#1}}

\renewcommand{\vec}[1]{\bm{\mathrm{#1}}}
\newcommand{\acro}[1]{\textsc{\MakeLowercase{#1}}}
\newcommand{\R}{\mathbb{R}}
\newcommand{\inv}{^{-1}}
\newcommand{\trans}{^{\top}}

\DeclareMathOperator*{\argmax}{arg\,max}

\begin{document}

\runningauthor{Henry Chai and Roman Garnett}

\twocolumn[
\aistatstitle{Improving Quadrature for Constrained Integrands}
\aistatsauthor{Henry Chai \\ hchai@wustl.edu \And  Roman Garnett \\ garnett@wustl.edu}
\aistatsaddress{Department of Computer Science and Engineering \\ Washington University in St. Louis}
]

\begin{abstract}

We present an improved Bayesian framework for performing inference of affine
transformations of constrained functions. We focus on quadrature with
nonnegative functions, a common task in Bayesian inference.  We consider
constraints on the range of the function of interest, such as nonnegativity or
boundedness. Although our framework is general, we derive explicit approximation
schemes for these constraints, and argue for the use of a log transformation
for functions with high dynamic range such as likelihood surfaces. We propose a
novel method for optimizing hyperparameters in this framework: we optimize the
marginal likelihood in the original space, as opposed to in the transformed
space. The result is a model that better explains the actual data. Experiments
on synthetic and real-world data demonstrate our framework achieves superior
estimates using less wall-clock time than existing Bayesian quadrature
procedures.

\end{abstract}



\section{Introduction}
Integrals over model (hyper)parameters are frequently encountered in Bayesian
inference. Model selection, for example, is a fundamental concern in the course
of scientific inquiry: which of several candidate models best explains an
observed dataset $\data$? The Bayesian approach requires the computation of
\emph{model evidence,} an integral of the form $Z = \int f(\data \given
\theta)\,\pi(\theta)\intd{\theta}$ where $\theta$ is a vector of model
parameters, $f(\data \mid \theta)$ is a likelihood, and $\pi(\theta)$ is a
prior. Computing a marginal predictive distribution similarly requires
integrating a predictive density $p(y \given x, \data, \theta)$ against a
posterior distribution $p(\theta \given \data)$. Note that the integrand in both
these scenarios is known \emph{a priori} to be nonnegative, as it is the product
of probability densities. Unfortunately, these integrals are often
computationally intractable and thus must be approximated.

Numerous common techniques to estimate such integrals rely on \emph{Monte Carlo}
estimators \citep{neal01, meng96, skilling04}. These methods are agnostic to
prior information about the integrand, such as nonnegativity, and also converge
slowly in terms of the number of required samples, rendering them ill-suited for
settings where the integrand is expensive to evaluate. One alternative is
\emph{Bayesian quadrature} (\acro{BQ}) \citep{larkin72, diaconis88, hagan91,
  rasmussen03}, which relies on a probabilistic belief on the integrand that can
be conditioned on observations to derive a posterior belief about the value of
the integral or any other affine transformation. The theoretical properties of
kernel quadrature methods (including \acro{BQ}) have been studied at length.
These methods can achieve faster convergence rates than Monte Carlo estimators
\citep{briol15, bach17, karvonen18}, even when the underlying model is
misspecified \citep{kanagawa16, kanagawa17}, a commonly-cited pitfall of
kernel-based methods.


Recent work by \citet{gunter14} and \citet{osborne12} have improved the speed
and accuracy of classical \acro{BQ} methods such as \emph{Bayesian Monte Carlo}
(\acro{BMC}) \citep{rasmussen03} for estimating integrals of \emph{nonnegative}
functions. These two methods reason about the square root and the log of the
integrand, respectively, instead of the integrand itself. By ``undoing'' these
transformations, we may softly incorporate the nonnegativity
constraint. Although previous work \citep{gunter14, osborne12} has demonstrated
that suitably modified \acro{BQ} can outperform Monte Carlo methods and
\acro{BMC} for estimating integrals of nonnegative functions, a general
framework for quadrature with the use of transformations has never been offered.

Our contribution is to define a Bayesian framework for a wide variety of
inference tasks, including quadrature, involving a broader class of constrained
functions. We provide complete details of this framework for two important
classes of constrained functions: nonnegative functions and functions bounded on
an interval. Common examples arising in machine learning include likelihoods and
classification (e.g., validation) error. We then apply our framework to
quadrature, where we address some shortcomings of previous work. Specifically,
our approach can make effective use of a log transform to efficiently estimate
integrals involving extreme dynamic range. This is in contrast to the method in
\citep{gunter14}, which cannot handle such dynamic range, and to
\citep{osborne12}, which relied on a series of abstruse and inefficient
approximations. Finally, we develop a novel training procedure whereby
hyperparameters are fit by maximizing the marginal likelihood of true
observations of the integrand. All previous related work instead fit
hyperparameters by maximizing the marginal likelihood of transformed
observations. We demonstrate this can lead to undesirable behavior and that our
procedure yields a better-behaved model, \emph{even if simply adopted into
  previous procedures such as} \citep{gunter14}.  We conduct experiments with
real-world data showing that our proposed framework and novel hyperparameter
optimization method outperforms previous \acro{BQ} algorithms.


\section{Bayesian Quadrature}
Let $Z = \int f(x)\,\pi(x)\intd{x}$ be an intractable integral.\footnote{\noindent 
For notational simplicity, the following will be written as if $x\in\R$, but all
results extend to $x\in\R^d$.}  Bayesian quadrature operates by placing a
Gaussian process (\acro{GP}) prior on the function $f$, $p(f) = \mc{GP}(\mu,
\Sigma)$ \citep{rasmussen06}. \acro{GP}s are probability distributions over
functions, where the joint distribution of any finite number of function values
is multivariate normal; this belief is parametrized by a mean function $\mu(x)$
and a covariance function $\Sigma(x, x')$. Given a set of observations at
locations $\vec{x}=\{x_1,\ldots,x_n\}$ with corresponding function values
$\vec{f} = f(\vec{x})$, the \acro{GP} prior can be conditioned on these
observations to arrive at a posterior \acro{GP} with mean
$\mu_\data(x)=\mu(x)+\Sigma(x,\vec{x})\Sigma(\vec{x},\vec{x})^{-1}\bigl(\vec{f}-
\mu(\vec{x})\bigr)$ and covariance
$\Sigma_\data(x,x')=\Sigma(x,x')-\Sigma(x,\vec{x})\Sigma(\vec{x},\vec{x})^{-1}
\Sigma(\vec{x},x')$.

Given a \acro{GP} belief on a function, we may derive a belief over integrals
of that function using the fact that \acro{GP}s are closed under linear
transformations such as integration \cite{rasmussen03}. Specifically, if $p(f) =
\mathcal{GP}(\mu,\Sigma)$, then our integral of interest $Z = \int
f(x)\,\pi(x)\intd x$ is normal:
\begin{equation}
  \textstyle
  p(Z)
  =
  \mathcal{N}\bigl(
  \int\mu(x)\pi(x)\intd{x},
  \iint \Sigma(x,x')\pi(x)\pi(x')\intd{x}\intd{x'}\bigr).
  \label{eq1}
\end{equation}

Warped sequential active Bayesian integration (\acro{WSABI}) \citep{gunter14}
builds off \acro{BQ} to incorporate nonnegativity information about an integrand
$f$ with a warped \acro{GP} \citep{snelson04}. Specifically, \acro{WSABI} places
a \acro{GP} prior on $g(x)=\sqrt{2(f(x)-\alpha)}$, for some small positive
constant $\alpha$. This prior is then conditioned on observations to arrive at a
posterior, like \acro{BQ}. Warped \acro{GP}s have been previously used for a
variety of machine learning tasks \citep{zhang10, schmidt09}. However, when
applied to quadrature, warped \acro{GP}s lack the key property of closure under
linear transformations. In particular, the marginal predictive distribution of
an arbitrary function value $f(x)$ is no longer Gaussian but instead depends on
the choice of warping function; in the case of \acro{WSABI,} these marginals are
non-central $\chi^2$ distributions, which are inconvenient for
quadrature. \acro{WSABI} approximates the posterior belief about $f$ as a
\acro{GP} using one of two proposed approximation schemes: linearization, which
uses a first-order Taylor expansion around the posterior mean of the \acro{GP}
on $g(x)$, and moment matching, which calculates the mean and covariance of the
true posterior distribution on $f$ and adopts a \acro{GP} matching these moments
\citep{gunter14}. Either approximation gives a \acro{GP} belief about $f$
approximately incorporating the nonnegativity constraint, and we may use
standard results such as \eqref{eq1} to reason about integrals, etc.  Below we
will describe a general procedure following these ideas, then describe how to
improve upon the procedure in numerous ways in practice.


\section{Inference on Constrained Functions}
\label{framework}

We propose a framework for inferring affine functionals of functions with
contraints on their range. Let $f\colon \mc{X} \to \mc{Y} \subset \R$ be a
function of interest with range constrained to a subset $\mc{Y}$ of the real
line; for example, a strictly positive function would have $\mc{Y} = (0,
\infty)$, and a function bounded on an interval would have $\mc{Y} = (a,
b)$. Let $Z = L[f]$ be an affine functional of $f$ we wish to infer.
\begin{enumerate}
\item
  Determine an invertible warping $\xi$ mapping $\R$ onto $\mc{Y}$, the domain
  of $f$. Define an \emph{unconstrained} function $g\colon \mc{X} \to \R$ by
  $g(x)=\xi^{-1}\bigl(f(x)\bigr)$ and place a \acro{GP} prior on $g$, $p(g) =
  \mathcal{GP}(\mu,\Sigma)$.
\item
  Observe $g$ at locations chosen by an appropriate sampling policy, yielding
  data $\data=\{\vec{x}, g(\vec{x})\}$.
\item
  Derive a posterior belief on the transformed function, $p(g \given \data) =
  \mathcal{GP}(\mu_\data,\Sigma_\data)$.
\item
  Calculate the posterior mean $m_\data$ and covariance $K_\data$ functions of
  the induced posterior belief on $f$. If needed, these can be approximated as
  polynomials in the posterior moments of $g$; see below for
  details. Approximate the belief on $f$ by a moment-matched \acro{GP}: $p(f
  \given \data) \approx \mathcal{GP}(m_\data, K_\data)$.
\item
  Derive a posterior belief about $Z$ (e.g., \eqref{eq1}):
  \begin{equation}
    p(Z \given \data) = \mathcal{N}\bigl(L[m_\data], L^2[K_\data]\bigr)\label{eq2}
  \end{equation}
	where $L^2[K]=L\bigl[L[K(x,\cdot)]\bigr]=L\bigl[L[K(\cdot,x)]\bigr]$ (see \eqref{eq1} for an example).
\end{enumerate}

\begin{table*}[t]
  \caption{Induced moments of $f = \xi(g)$ for various transformations $\xi$, if
    $p(g) = \mc{GP}(\mu, \Sigma)$. We provide the \emph{raw} second moment $C(x,
    x')$ in this table; the covariance function can be computed by $K(x, x') =
    C(x, x') - m(x)\,m(x')$. Some entries for the second raw moment refer to
    values of the first moment for that transform.}
  \label{tab1}
  \centering
  \begin{small}
    \begin{tabular}{lcc}
      \toprule
      transform
      & \multicolumn{1}{c}{first moment $m(x) = \mathbb{E}\bigl[f(x)\bigr]$}
      & \multicolumn{1}{c}{second raw moment $C(x, x') = \mathbb{E}\bigl[f(x)f(x')\bigr]$} \\
      \midrule
      $\xi(f) = \alpha + f^2$ \citep{gunter14} &
      $\alpha + \mu(x)^2 + \Sigma(x, x)$ &
      $2\Sigma(x, x')^2 + 4\mu(x)\,\Sigma(x, x')\,\mu(x') + m(x)\,m(x')$
      \\
      $\xi(f) = \text{any polynomial in $f$}$ &
      polynomial in $\mu$ and $\Sigma$ &
      polynomial in $\mu$ and $\Sigma$
      \\
      $\xi(f)=\exp(f)$ &
      $\exp\bigl(\mu(x)+\nicefrac{1}{2}\Sigma(x,x)\bigr)$ &
      $m(x)\, \exp\bigl(\Sigma(x, x')\bigr)\, m(x')$ \\
      $\xi(f)=\Phi(f)$ &
      $\displaystyle \Phi\biggl(\frac{\mu(x)}{\sqrt{\Sigma(x,x)+1}}\biggr)$ &
      $\Phi\left(\begin{bmatrix}\mu(x) \\ \mu(x')\end{bmatrix}, \begin{bmatrix}\Sigma(x,x)+1 & \Sigma(x,x') \\ \Sigma(x',x) & \Sigma(x',x')+1\end{bmatrix}\right)$ \\
      \bottomrule
    \end{tabular}
  \end{small}
\end{table*}

In short, we maintain a \acro{GP} belief on a warped version of $f$ that removes
the constraint. We then approximate a \acro{GP} belief on $f$ given data via
moment matching, after which we can easily reason about affine functionals.
Particular instances of this framework have appeared in the literature; for
example, \acro{WSABI} (specifically the \acro{--M} variant \citep{gunter14})
implements this framework using the square root transform to infer integrals of
nonnegative functions. However, we will discuss the framework in greater
generality and provide practical advice.

The above framework is agnostic to several design choices. First, we do not
specify the warping function $\xi$ in step (1). \acro{WSABI}, for example,
relies intimately on the square root map. This induces nonnegativity, but we
will demonstrate that it does not yield useful models for functions with high
dynamic range. We will provide details to work with a wide range of warping
functions, including polynomials, log transformations, and sigmoidal
transformations such as the probit.

Further, we do not specify how exactly the posterior belief in the transformed
space $p(g \given \data)$ is derived in step (3), in particular how any
associated hyperparameters are fit. We will discuss this issue in detail later
and provide a novel approach.

Finally, we make no assumptions about the mechanism for choosing observation
locations $\vec{x}$ in step (2). These could be sampled proportional to some
distribution, \`{a} la Monte Carlo, or chosen via information-theoretic
principles or some other scheme. If no warping function is used, as in
\acro{BMC,} then the optimal set of locations in terms of minimizing the
posterior variance/entropy of our belief about $Z$ can be precomputed, as the
posterior covariance of a \acro{GP} does not depend on the observed values
\citep{minka00}. However, in the scheme outlined above, the approximate
posterior covariance of $f$, $K_\data$, \emph{does} depend on the observed
values, as it a function of the mean belief in the transformed space,
$\mu_\data$; see below for details. Thus, to make use of policies that maximize
information gain in this setting, observation locations must be selected
sequentially. In \acro{WSABI}, samples are chosen by greedily maximizing
information gain about the integrand, selecting each point to maximize the
posterior variance: $x^*=\argmax_x K_\data(x, x)$. \citet{osborne12} chose
samples so as to maximize the expected information gain about an integral $Z$
directly. Both are compatible with our proposed framework.


\subsection{Transform selection, moment matching}
\label{moments}
We briefly pause to discuss the moment-matching step in step (4) of our
procedure. Several useful general-purpose transformations admit closed-form
expressions for the posterior mean and covariance on $f$ given a \acro{GP}
belief about $g = \xi^{-1}(f)$, $p(g) = \mc{GP}(\mu, \Sigma)$. We provide a
summary for several notable examples in Table \ref{tab1}; details can be found
in the supplemental material.

\begin{figure}
  \hspace*{0.25cm}
  \includegraphics{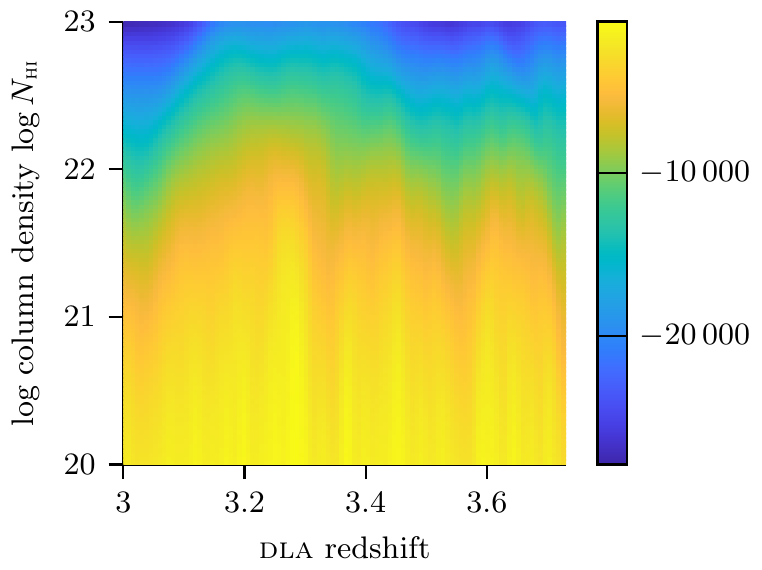}
  \caption{The log-likelihood surface for a real-world astronomical dataset
    corresponding to a an astronomical model described further in our
    experiments \citep{garnett17}. The dynamic range is massive, on the order of
    $\exp(27\,135) \gg 10^{10\,000}$.}
  \label{dla_dynamic_range}
\end{figure}

For a strictly positive function taking values on $\mc{Y} = (0, \infty)$, we may
use the square root transform $\xi^{-1} = \sqrt{f}$ or the log transform
$\xi^{-1} = \log f$.  Choosing an appropriate transform for a given scenario
will require consideration of the data. For example, when the data has extreme
dynamic range, as is often the case for likelihood surfaces, a log
transformation may be desired. Figure \ref{dla_dynamic_range} shows an example
\emph{log} likelihood surface for a real-wold astronomical model we will
consider in our experiments \citep{garnett17}. Note that computing a model
evidence requires integrating the \emph{likelihood} surface, not the log
likelihood. The dynamic range of the likelihood is on the rough order of
$10^{10\,000}$, and no off-the-shelf \acro{GP} could reasonably model this
function. The square root of the likelihood, as would be used in \acro{WSABI},
reduces the dynamic range to an equally unmanageable $10^{5000}$. The log
transformation, however, produces a well-behaved surface that could be
reasonably modeled with a standard \acro{GP}.

To model a bounded function taking values on the interval $(0,1)$, we could use
a probit transform $\xi = \Phi(f);$ closed-form moments for the induced belief
on $f$ are also provided. The covariance requires the bivariate Gaussian
\acro{CDF}, which can be estimated efficiently with high precision
\citep{genz04}. By shifting and scaling appropriately, we can model a function
taking values on any interval of the form $(-\infty, a)$,$(a, b)$, or $(b,
\infty)$.

For an arbitrary polynomial warping $\xi = a_n f^n + a_{n - 1}f^{n-1} \cdots +
a_0$, an extension of Isserlis' theorem guarantees that the moments of $f$ will
be polynomials in $\mu$ and $\Sigma$ (of degree $n$ for the mean and $2n$ for
the covariance), and a simple algorithm can generate these moments on demand
\citep{withers85}.

\begin{figure}
  \hspace*{1.85cm}no transform\\
  \hspace*{0.5cm}%
  \includegraphics{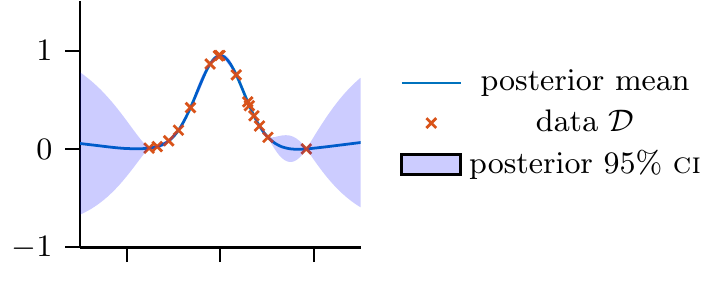}\\
  \begin{tabular}{cc}
    \hspace*{1.15cm}log transform &
    \hspace*{-0.5cm}\hspace*{0.8cm}probit transform
    \\[1ex]
    \includegraphics{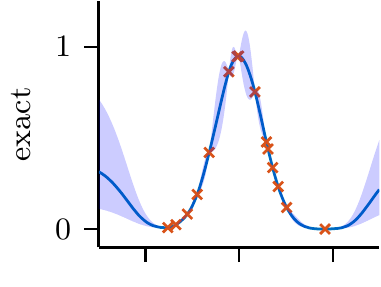} &
    \hspace*{-0.5cm}
    \includegraphics{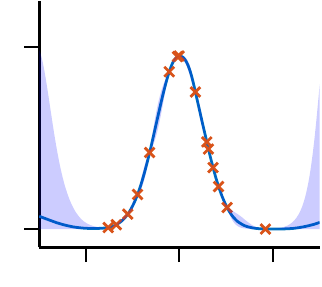}
    \\
    \includegraphics{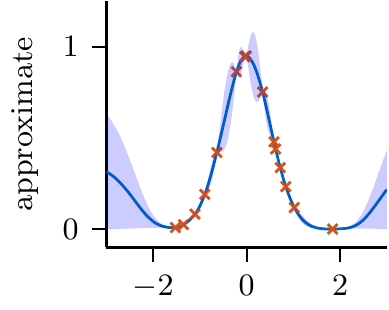} &
    \hspace*{-0.5cm}
    \includegraphics{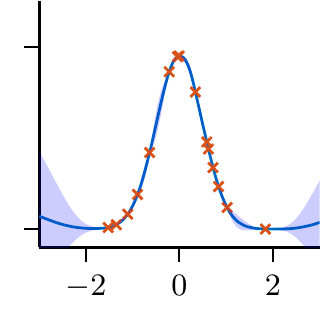}
    \\
  \end{tabular}
  \caption{A demonstration of fitting a simple function $f(x) = 0.95
    \exp(-2x^2)$ on the interval $[-3, 3]$ using a log and probit transformation
    in our framework. Each column shares an $x$ axis and each row shares a $y$
    axis.}
  \label{transform_demo}
\end{figure}

We show a brief demonstration of fitting the bounded function $f(x) =
0.95\exp(-2x^2)$ (scaled to avoid the value of exactly 1 at 0) using a log and
probit transformation in Figure \ref{transform_demo}. The model fit to data
directly and unaware of the transformation produces considerable predictive mass
on invalid values. The exact posteriors for the log and probit transformations
both absolutely respect their respective constraints. The moment-matched
\acro{GP}s are excellent approximations.

\begin{figure*}
\begin{subfigure}[b]{0.24\textwidth}
  \centering
  \includegraphics{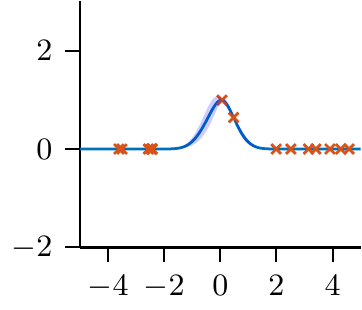}
  \caption{$p(f)$, fit in $f$-space}
  \label{fig_2a}
\end{subfigure}
\begin{subfigure}[b]{0.24\textwidth}
  \centering
  \includegraphics{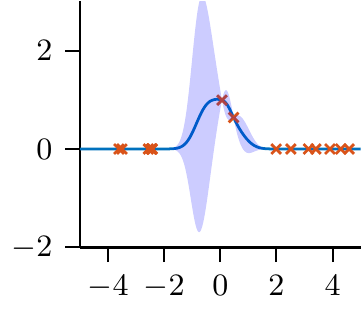}
  \caption{$p(f)$, fit in $g$-space}
  \label{fig_2b}
\end{subfigure}
\begin{subfigure}[b]{0.24\textwidth}
  \includegraphics{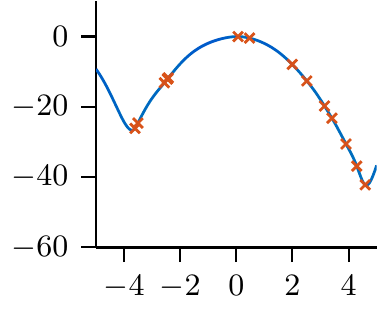}
  \caption{$p(g)$, fit in $f$-space}
  \label{fig_2c}
\end{subfigure}
\begin{subfigure}[b]{0.24\textwidth}
  \includegraphics{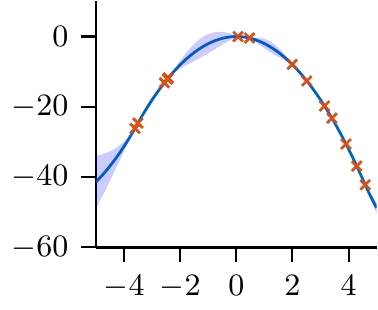}
  \caption{$p(g)$, fit in $g$-space}
  \label{fig_2d}
\end{subfigure}
\caption{Fitting in $f$-space vs.\ fitting in $g$-space. We model the function
  $f(x)=0.95\exp(-2x^2)$ on the interval $[-5, 5]$, conditioning on 15 observations
  at locations sampled uniformly at random.  We place a \acro{GP} prior on
  $g=\log f$ with constant mean and Mat\'ern covariance with $\nu=\nicefrac{3}{2}$.
  This model has three hyperparameters: a mean, an output scale, and a length scale.
  These were fit in $f$-space ((a) and (c)) and $g$-space ((b) and (d)). See the
  legend in Figure \ref{transform_demo}.}
\label{fig2}
\end{figure*}

\subsection{Hyperparameter optimization}
\label{hyp_opts}
When \acro{GP}s are used for inference, an important consideration is how to set
the associated hyperparameters. One commonly used method is to optimize the
marginal likelihood of the observed data using gradient-based methods as the
gradient of the marginal likelihood w.r.t.\ hyperparameters is readily available
for this model class. The motivation for fitting hyperparameters by maximizing
the marginal likelihood is to explain the observed data as well as
possible. However, when performing inference using the above framework, the goal
is not to have the best possible explanation of the \emph{transformed} data, but
rather to have an accurate belief about the \emph{original, untransformed}
data. Previous related approaches (e.g., \citep{osborne12, gunter14}) have
ignored this fact and fit the hyperparameters of the warped \acro{GP} in the
warped space. We will show this can lead to poor behavior.

We propose setting hyperparameters by maximizing the marginal likelihood of the
untransformed data using the (approximate) posterior belief on $f$; we will
refer to optimizing the hyperparameters in this manner as ``fitting in
$f$-space'' as opposed to ``fitting in $g$-space.''

Formally, if $p(g) = \mathcal{GP}\bigl(\mu(\theta),\Sigma(\theta)\bigr)$ (where
dependence on hyperparameters $\theta$ has been written explicitly), our
framework approximates $p(f)$ with $p(f) \approx
\mathcal{GP}\bigl(m\bigl(\mu(\theta),\Sigma
(\theta)\bigr),K\bigl(\mu(\theta),\Sigma(\theta)\bigr)\bigr)$. The exact
relationship between $\theta$ and the mean/covariance of $f$ depends on the
transformation $\xi$. For many natural choices, the partial derivatives
$\nicefrac {\partial m}{\partial\mu}$, $\nicefrac{\partial m}{\partial\Sigma}$,
$\nicefrac{\partial K}{\partial\mu}$ and $\nicefrac{\partial K}{\partial\Sigma}$
will be available.  Thus, we can evaluate the partial derivative of $f$
w.r.t.\ to $\theta$ and use the same gradient-based methods used to fit
hyperparameters in $g$-space to fit hyperparameters in $f$-space; for the
transformations found in Table \ref{tab1}, the relevant partial derivatives can
be found in the supplementary material.

Figure \ref{fig2} shows the impact of fitting the hyperparameters in $f$-space
as opposed to fitting in $g$-space using our toy function $f(x)=0.95\exp(-2x^2)$.
The hyperparameters learned in $f$-space result in a model that fits the
$f$-space data well but do a poor job explaining the data in $g$-space; the
learned mean is much higher than the mean of the transformed data and the
learned output scale is very small, leading to unreasonably little uncertainty
in the model. However, these learned hyperparameters make sense in the context
of the $f$-space data, where most of the observations are effectively zero and
the maximum observed value is slightly less than one. Conversely, the
hyperparameters learned in $g$-space fit the $g$-space data very cleanly, with a
well-scaled uncertainty. However, this translates to a poorly-behaved model in
$f$-space; the region from $[-2,-0]$ has what appears to be a very reasonable
variance in $g$-space, but this corresponds to a massive variance in $f$-space
that strongly defies the nonnegativity constraint.

We offer two practical notes about fitting in $f$-space in the case of a log
transform learned through our experiments. First, we suggest shifting the
$g$-space data so that the maximum observed value is exactly zero, as this
places the observations into a regime where the inverse transformation is
well-behaved. We are free to make such a shift as doing so simply scales the
$f$-space data by a constant.  Second, initializing the hyperparameter
optimization procedure must be done carefully when fitting in $f$-space. If one
is using a constant mean, we recommend avoiding na\"{i}vely initializing the
prior mean to be the mean of the transformed data.  Instead, we initialized the
mean to one of $-1$, $-2$, $-5$, and $-10$ and initialized the output scale of
the covariance function to the mean initialization divided by $-2$.  We believe
this set of initializations to be sufficient after shifting the data because the
relevant portions of the $f$-space data should be well-described by a
hyperparameter setting reachable from these initializations. Lower means may
result in undesirable behavior, as the corresponding output scales would need to
be large to explain the shifted observation at zero.

\subsection{Approximating the posterior on $Z$}
\label{taylor}
For some combinations of linear functionals and warping functions, the posterior
belief on $Z$ \eqref{eq2}, may be intractable, i.e., either $L[m]$ or $L^2[K]$
cannot be expressed in closed form. This is the case for quadrature with the log
transformation and most common choices of covariance function, including the
Mat\'ern and squared exponential kernels, as the posterior belief contains a
term of the form $\int\exp \exp x \intd{x}$.

Various approximation techniques can be used to estimate these intractable
quantities. \citet{osborne12} use \acro{BQ} itself, a somewhat unsatisfying
approach as it leads to infinite regress. \citet{briol15} provide a theoretical
justification for the use of Monte Carlo based methods when estimating
intractable posterior means. We propose an alternative approximation scheme that
makes use of a Taylor series expansion to approximate the $f$-space moments
$m(x)$ and $K(x, x')$. The exact nature of the Taylor series will depend on the
warping function $\xi$; for $\xi =\exp f$, the following approximations follow
from the expressions in Table \ref{tab1}:
\begin{align}
  m(x) &\approx 1+\mu(x)+\nicefrac{1}{2}\Sigma(x,x)\nonumber\\
  &\qquad\mspace{-3mu}+\bigl(\mu(x)+\nicefrac{1}{2}\Sigma(x,x)\bigr)^2/2+\dots\label{eq5}\\
  K(x,x') &\approx 1+\Sigma(x,x')+\nicefrac{1}{2}\Sigma(x,x')^2\nonumber\\
  &\qquad\mspace{-3mu}+\Sigma(x,x')\bigl(\mu(x')+\nicefrac{1}{2}\Sigma(x',x')\nonumber\\
  &\qquad\mspace{-3mu}+\mu(x)+\nicefrac{1}{2}\Sigma(x,x)\bigr)+\dots\label{eq6}
\end{align}
Given these approximations, the posterior mean and variance for quadrature are
tractable for certain covariance functions, including the squared exponential
kernel \citep{hennig16}. Indeed, for reasonably well-behaved warpings $\xi$, the
associated approximations will be polynomial functions of $\mu$ and $\Sigma$,
and thus tractable for integrating against standard covariance functions.  This
last result follows directly from Isserlis' theorem (see \S\,\ref{moments}).
Unfortunately, computing this approximation is expensive for higher-order terms:
computing the $d$th order term in either Taylor series after making $n$ function
evaluations takes $\Theta(n^{2d})$ time.


\section{Experiments}
\label{exps}
We perform experiments in a variety of settings to evaluate our proposed
framework and demonstrate the importance of our proposed improvements. We begin
by exploring the effect of fitting in $f$-space using different transformations
on a simple regression task. Then we apply our framework to quadrature of
nonnegative integrands using a moment-matched log transformation
(\acro{MMLT}). We compare these results against \acro{WSABI} and \acro{BMC} as
well as Monte Carlo methods. If not otherwise specified, all \acro{GP} priors
were chosen to have constant mean and Mat\'ern covariance with
$\nu=\nicefrac{3}{2}$, all sample locations were selected iteratively using
uncertainty sampling in $f$-space \citep{gunter14}, all hyperparameters were fit
in $f$-space when applicable, and all intractable integrals were estimated
using quasi-Monte Carlo \citep{caflisch98}.

\subsection{Hyperparameter tuning}
\label{reg}
To assess the impact of modeling constrained functions using transformations, we
consider three regression tasks using the standard benchmarks of the
\acro{HPO}lib package \citep{eggensperger13}: online \acro{LDA}, \acro{SVM}, and
logistic regression (\acro{LR}). For each benchmark, \citet{eggensperger13}
provide a list of hyperparameter settings for the eponymous machine learning
algorithm along with the associated observations of some relevant, machine
learning quantity: for the online \acro{LDA} benchmark, the observed values are
per-word perplexities (which are nonnegative), whereas for the \acro{SVM} and
\acro{LR} benchmarks the observed values are prediction error rates
(which are bounded between 0 and 1). The online \acro{LDA}, \acro{SVM}, and
\acro{LR} datasets contain 289, 1400, and 9680 observations, respectively.

For each benchmark, we ran the following experiment 100 times: we randomly
select some percentage of the dataset to be a training set (20\% for online
\acro{LDA}, 5\% for the other two) and designate the remaining observations to
be a test set. We fit a moment-matched \acro{GP} to the training set using both
the log and square root transformations for online \acro{LDA} and a probit
transformation for both \acro{SVM} and \acro{LR}. We compare our framework
against a standard, constraint-unaware \acro{GP} and a moment-matched \acro{GP}
where the hyperparameters were fit in $g$-space as opposed to in $f$-space. We
consider two metrics: the root mean squared error (\acro{RMSE}) on the test set
and the mean predictive log likelihood (\acro{MLL}) of observations in the test
set, $\E\bigl[\log p\bigl(f(x) \mid x, \data\bigr)\bigr]$.

\begin{table}
  \caption{Regression experiment results.}
  \label{tab2}
  \centering
  \begin{small}
    \begin{tabular}{crcc}
      \toprule
      dataset & \multicolumn{1}{c}{transform} & \acro{RMSE} & \acro{MLL} \\
      \midrule
      \multirow{5}{*}{\acro{LDA}} & none & $153$ & $-1.0 \times 10^{10}$ \\
      & square root ($g$-space) & $142$ & $-2.1 \times 10^{6\phantom{0}}$ \\
      & square root ($f$-space) & $142$ & $-6.1 \times 10^{5\phantom{0}}$ \\
      & log ($g$-space) & $134$ & $-4.1 \times 10^{6\phantom{0}}$ \\
      & log ($f$-space) & $133$ & $-4.8 \times 10^{5\phantom{0}}$ \\ \midrule
      \multirow{3}{*}{\acro{SVM}} & none & $0.015$ & $\phantom{-}2.83$ \\
      & probit ($g$-space) & $0.015$ & $\phantom{-}2.82$ \\
      & probit ($f$-space) & $0.015$ & $\phantom{-}2.91$ \\ \midrule
      \multirow{3}{*}{\acro{LR}} & none & $0.036$ & $\phantom{-}1.98$ \\
      & probit ($g$-space) & $0.036$ & $\phantom{-}2.06$ \\
      & probit ($f$-space) & $0.035$ & $\phantom{-}2.07$ \\ \midrule
      \multirow{3}{*}{\acro{IM}} & none & $0.281$ & $-0.110$ \\
      & probit ($g$-space) & $0.266$ & $-0.324$ \\
      & probit ($f$-space) & $0.256$ & $\phantom{-}0.319$ \\
      \bottomrule
    \end{tabular}
  \end{small}
\end{table}

The results are shown in Table \ref{tab2}. We can extract a few trends.  Using a
transformation that respects the \emph{a priori} knowledge about the target
function leads to an improvement in accuracy; for the online \acro{LDA}
benchmark, the difference between the \acro{RMSE} of the constraint-agnostic
\acro{GP} and the \acro{RMSE}s of all methods using a transformation is
significant at the 1\% significance level according to a one-sided paired
$t$-test. In general, our proposed hyperparameter optimization methodology does
not lead to a significant difference in the \acro{RMSE}. All methods tend to
learn similar predictive means in $f$-space for these datasets, which do not
reflect extreme behavior. The impact of our proposed methodology can be seen in
the mean predictive log likelihoods, however. In terms of this metric, fitting
in $f$-space is preferable to fitting in $g$-space for both transforms as it
leads to better-scaled uncertainties.

The gains of fitting in $f$-space are reduced when using the probit
transformation on these particular benchmarks because the dynamic range is not
very large: observations of the per-word perplexity in the \acro{LDA} benchmark
range from roughly 1000 to 5000, whereas observations of the error rates for the
\acro{SVM} and \acro{LR} benchmarks only range from 0.24 to 0.50 and
from 0.07 to 0.91, respectively. Although the range of observations for the
\acro{LR} benchmark may seem large, this translates to observations
between $-1.5$ and 1.5 in the transformed space.

To showcase the power of the probit transformation with more-extreme data, we
ran the following in-model (\acro{IM}) experiment 100 times. We randomly sampled
a draw from a two-dimensional \acro{GP} prior, which we then pushed through the
inverse-probit transformation to generate a function bounded between 0 and
1. The output scale and length scales of the \acro{GP} were set such that
samples range roughly from $-5$ to 5 over the domain. We then sampled 200 points
from the draw, fit a moment-matched \acro{GP} using the probit transform (in
both $f$-space and $g$-space) to 20\% of the points, and predicted the values of
the remaining 80\%. The results are shown in Table \ref{tab2}. All differences in
performance are significant at the 1\% significance level according to one-sided
paired $t$-tests. As the results indicate, in this setting, it becomes important
to fit hyperparameters in $f$-space rather than in $g$-space to achieve
reasonably scaled uncertainties.

\subsection{Detecting DLAs via model selection}
\label{dla_exps}

We consider a real-world quadrature application of our framework, a model
selection problem from astrophysics. We wish to infer whether a damped
Lyman-$\alpha$ absorber (\acro{DLA}) exists along the line of sight between a
quasar and earth given spectrographic observations. \acro{DLA}s are large
gaseous clouds containing neutral hydrogen at high densities.  Their location
and size can be inferred from observations of quasar spectra as they cause
distinctive dips in the observed flux at well-defined wavelengths.  The
distribution of \acro{DLA}s throughout the universe is important as it provides
insight into models of galaxy formation. \citet{garnett17} developed a model
that specifies the likelihood that a given emission spectrum contains a putative
\acro{DLA}. The model is parameterized by two physical features of a candidate
\acro{DLA}: its column density, which roughly corresponds to its size, and its
redshift, which roughly corresponds to its distance from earth. \citet{garnett17}
also specified a data-driven prior distribution over these two parameters, which
must be integrated against to calculate the model evidence and derive a posterior
distribution of \acro{DLA} presence. The model evidence of this \acro{DLA} model
is an (intractable) integral of the likelihood over the domain of these two model
parameters. Here we will consider computing the model evidence of 2000 spectra
gathered from phase \acro{III} of the Sloan Digital Sky Survey (\acro{SDSS--III})
\citep{eisenstein11}. For a complete description of the problem, data, and model,
see \citep{garnett17}.

A sample log-likelihood surface for this model corresponding to a particular
quasar spectrum is shown in Figure \ref{dla_dynamic_range}. These functions are
highly multimodal and have a massive dynamic range. These features make
computing the model evidence a difficult task for alternative methods such as
\acro{BMC} and \acro{WSABI}. One convenient feature of this experimental setting
is that the dimensionality of the intractable integral can be scaled up to any
even number simply by calculating the model evidence for the existence of $n$
\acro{DLA}s, resulting in a 2$n$-dimensional integral \citep{garnett17}.

\begin{figure*}[!t]
\captionsetup[subfigure]{labelformat=empty}
\begin{subfigure}{0.5\textwidth}
    \centering
    \includegraphics{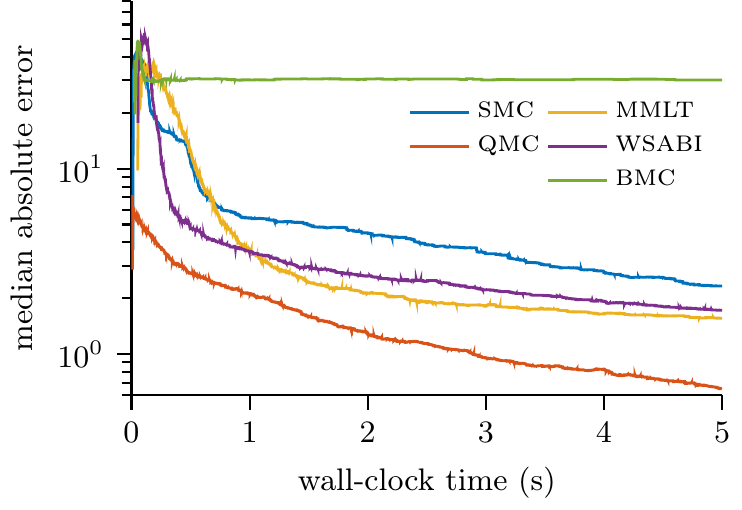}
    \caption{(a) 2d model evidence}
    \label{fig_4a}
\end{subfigure}
\begin{subfigure}{0.5\textwidth}
    \centering
    \includegraphics{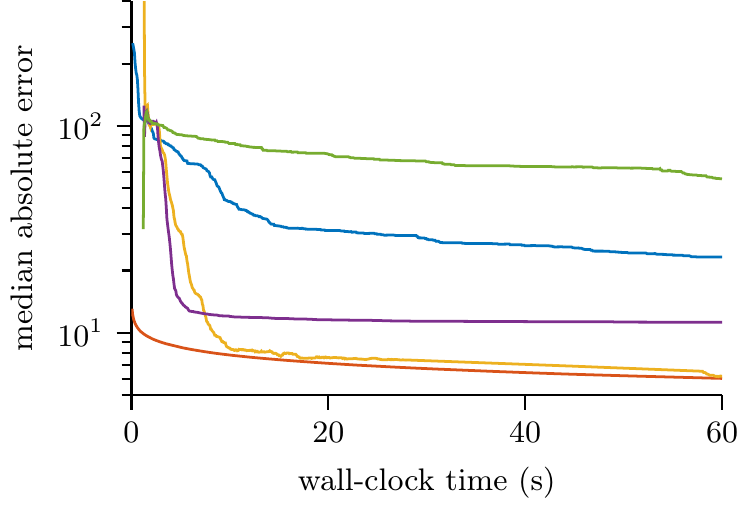}
    \caption{(b) 6d model evidence}
    \label{fig_4b}
\end{subfigure}
\caption{The median absolute predictive error of each method's estimate of the
  log model evidence over time in the \acro{DLA} experiments.
}
\label{fig4}
\end{figure*}

We conducted an experiment comparing the accuracy of \acro{BQ} methods for
estimating model evidence in this setting, including \acro{BMC}, \acro{WSABI},
and \acro{MMLT}. We considered the latter two fitting both in $f$-space and in
$g$-space.  We also compared with sequential Monte Carlo (\acro{SMC}) and
quasi-Monte Carlo (\acro{QMC}) estimation. We estimate model evidences for a single
\acro{DLA} and three \acro{DLA}s in 2000 quasar spectra, entailing two- and
six-dimensional integrals, respectively.  Each method was allotted 5 seconds of
wall-clock time for estimating the two-dimensional integrals and 60 seconds for
the six-dimensional integrals. Monte Carlo methods drew or constructed samples
from the prior specified by \citet{garnett17}.

\begin{table}
\caption{Mean $\log p(Z^*\given \data)$ at termination.}
\label{tab3}
\centering
\begin{small}
    \begin{tabular}{rcc}
    \toprule
    transform & 2d & 6d \\
    \midrule
    none (\acro{BMC}) & $-0.79$ & $\phantom{-}1.93$ \\
    square root (\acro{WSABI}) ($g$-space) & $\phantom{-}3.67$ & $\phantom{-}3.40$ \\
    square root (\acro{WSABI}) ($f$-space) & $\phantom{-}3.89$ & $\phantom{-}3.43$ \\
    log (\acro{MMLT}) ($g$-space) & $-266$ & $-505$ \\
    log (\acro{MMLT}) ($f$-space) & $\phantom{-}10.3$ & $\phantom{-}7.57$ \\
    \bottomrule
    \end{tabular}
\end{small}
\end{table}

\begin{table}
\caption{Mean \acro{MLL} at termination.}
\label{tab4}
\centering
\begin{small}
    \begin{tabular}{rcc}
    \toprule
    transform & 2d & 6d \\
    \midrule
    none (\acro{BMC}) & $-1.66$ & $\phantom{-}0.33$ \\
    square root (\acro{WSABI}) ($g$-space) & $\phantom{-}1.51$ & $\phantom{-}1.26$ \\
    square root (\acro{WSABI}) ($f$-space) & $\phantom{-}1.59$ & $\phantom{-}1.51$ \\
    log (\acro{MMLT}) ($g$-space) & $-3.87$ & $-7.28$ \\
    log (\acro{MMLT}) ($f$-space) & $\phantom{-}1.68$ & $\phantom{-}1.65$ \\
    \bottomrule
    \end{tabular}
\end{small}
\end{table}


Figure \ref{fig4} shows the median absolute error over time of each method,
using exhaustive \acro{QMC} sampling as ground truth. \acro{MMLT} outperforms
all other methods except \acro{QMC}; note that \acro{QMC} is not necessarily
well-suited for model-selection when it is not possible to construct an
appropriate low-discrepancy sequence, but we use it to provide a gold-standard
baseline. The difference in absolute errors at termination between \acro{MMLT}
and the other \acro{BQ} methods is significant for the six-dimensional integrals
at a 1\% significance level according to a one-sided paired $t$-test.



Tables \ref{tab3} and \ref{tab4} show the results of additional experiments performed
in this setting that demonstrate the importance of our proposed hyperparameter
optimization methodology. Table \ref{tab3} compares the log-likelihood of the true
value of the integral $Z^*$ under each Bayesian method's posterior belief upon
termination in these experiments while Table \ref{tab4} compares the \acro{MLL}
(see \S\,\ref{reg}). Here the \acro{MLL} is computed by averaging over the log predictive
probabilities of the \acro{QMC} samples used to estimate the model evidence.

\acro{MMLT} where the hyperparameters are fit in $f$-space outperforms all alternatives
on both metrics in both the two-dimensional and six-dimensional experiments; the
differences in Table \ref{tab3} are significant at a 1\% significance level according
to a one-sided paired $t$-test. \acro{MMLT} where the hyperparameters are fit in
$g$-space significantly \emph{underperforms} the other Bayesian algorithms. The
relatively poor performance of fitting in $g$-space on these metrics is largely due to
the high dynamic range of the likelihood surfaces, which forces the output scales learned
by fitting in $g$-space to be high. This in turn causes both the pointwise distributions
and the distribution on the value of the integral to have large variances (relative to
their means), making the likelihood everywhere low, much like the situation depicted in
Figure \ref{fig2}.

The difference between \acro{WSABI} where the hyperparameters are fit in $f$-space and
\acro{WSABI} where the hyperparameters are fit in $g$-space on both metrics is
relatively small. This is a consequence of the square root transformation, which barely
affects the extreme dynamic range of this data. The likelihood is so extremely small
everywhere (on the order of $10^{-10\,000}$) that there is practically no difference
between the true values and their square root. Thus, the settings of the hyperparameters
arrived at under the two methodologies are very similar; importantly, they have similar
output scales, thus explaining their similar uncertainties about both $f$ and $Z^*$.
However, for \acro{MMLT}, where the transformation does result in a drastic change in the
dynamic range of the observations, fitting in $f$-space is crucial as it ensures that all
the benefits of making this more useful transformation can be reaped. Nonetheless,
fitting hyperparameters in $f$-space in general will not decrease performance and can
result in significant gains.

\section{Conclusion}
\label{conc}
We have presented a general Bayesian framework for performing inference about
affine transformations of constrained functions. We developed a novel procedure
for optimizing the hyperparameters associated with our method whereby the
hyperparameters are set to maximize the marginal likelihood of the true data as
opposed to the transformed data. Although maximizing the marginal likelihood of
the transformed data may seem intuitive, we show that doing so can lead to
undesirable behavior, particularly if the target function has a wide dynamic
range. We then applied our proposed framework to perform regression on bounded
functions and both regression and quadrature on nonnegative functions. This novel
\acro{BQ} algorithm outperforms previously proposed algorithms on synthetic and
real-world data, both in terms of accuracy and speed of convergence. 



\section{Acknowledgements}
\label{ack}
This work was supported by the National Science Foundation under Award Number \acro{IIA}--1355406.


\newpage

\twocolumn[
\aistatstitle{Supplementary Materials for Improving Quadrature for Constrained Integrands}
]


\section{Derivation of Moments}
\label{math}

\begin{table*}[t]
  \caption{Induced (\emph{raw}) first and second moments of $f = \xi(g)$ for the
  log and probit transformations; the covariance function can be computed by
  $K(x,x') = C(x,x')-m(x)\,m(x')$. Some entries for the second raw moment refer
  to values of the first moment for that transform.}
  \label{tab1}
  \centering
  \begin{small}
    \begin{tabular}{lcc}
      \toprule
      transform
      & \multicolumn{1}{c}{first moment $m(x) = \mathbb{E}\bigl[f(x)\bigr]$}
      & \multicolumn{1}{c}{second raw moment $C(x, x') = \mathbb{E}\bigl[f(x)f(x')\bigr]$} \\
      \midrule
      $\xi(f)=\exp(f)$ &
      $\exp\bigl(\mu(x)+\nicefrac{1}{2}\Sigma(x,x)\bigr)$ &
      $m(x)\, \exp\bigl(\Sigma(x, x')\bigr)\, m(x')$ \\
			[1.5ex]
      $\xi(f)=\Phi(f)$ &
      $\displaystyle \Phi\biggl(\frac{\mu(x)}{\sqrt{\Sigma(x,x)+1}}\biggr)$ &
      $\Phi\left(\begin{bmatrix}\mu(x) \\ \mu(x')\end{bmatrix}, \begin{bmatrix}\Sigma(x,x)+1 & \Sigma(x,x') \\ \Sigma(x',x) & \Sigma(x',x')+1\end{bmatrix}\right)$ \\
      \bottomrule
    \end{tabular}
  \end{small}
\end{table*}

This section provides the derivation of the first and second raw moments
for the log and probit transforms, as shown in Table \ref{tab1}, and the relevant partial
derivatives, which are required to use gradient based methods to optimize the \acro{GP}
hyperparameters in $f$-space as described in the main text.

\subsection{Log transform moments}
Let $\vec{y}=\{y_1,\ldots,y_n\}$ be a multivariate Gaussian random variable with mean vector
$\vec{\mu}$ and covariance matrix $\Sigma$ and let $\vec{x}=\exp(\vec{y})$. Then $\vec{x}$
follows a multivariate log-normal distribution \citep{tarmast01, klugman12, halliwell15}, a
well-studied distribution whose first and second raw moments are given by
\begin{align}
	\E&[x_i]=\exp\left(\mu_i+\frac{\Sigma_{ii}}{2}\right) \\
	\E&[x_i^2]=\exp\left(2\mu_i+2\Sigma_{ii}\right) \\
	\E&[x_ix_j]=\exp\left(\mu_i+\mu_j+\frac{1}{2}(\Sigma_{ii}+\Sigma_{jj})+\Sigma_{ij}\right),
	\label{eq}
\end{align}
where $x_i$ is the $i^{\textrm{th}}$ element of the vector $\vec{x}$, $\mu_i$ is the mean of the
$i^{\textrm{th}}$ element of $y_i$ and $\Sigma_{ij}$ is the covariance between $y_i$ and $y_j$.
The derivation of these moments is omitted as they are well established in the literature and not
very interesting (they follow from a simple substitution and then completing the square within
the exponent).

In order to fit hyperparameters in $f$-space as described in the main text, we maximize
the likelihood of some observed training dataset (or equivalently, minimize the negative
log-likelihood) w.r.t. the hyperparameters of the \acro{GP} prior on the $g$-space belief.
Making use of equation 5.8 from \citet{rasmussen06}, it follows that the relevant quantities
are $\nicefrac{\partial \E[x_i]}{\partial\theta}$ and $\nicefrac{\partial \E[x_ix_j]}
{\partial\theta}$ where $\theta$ is some hyperparameter of either the mean or covariance
function of the \acro{GP} prior. Because the partial derivatives $\nicefrac{\partial\mu}
{\partial\theta}$ and $\nicefrac{\partial\Sigma}{\partial\theta}$ depend on the choice of mean
and covariance function, we instead present the partial derivatives of the moments w.r.t. the
means and covariances/variances. These partial derivatives can be used in conjuction with
$\nicefrac{\partial\mu}{\partial\theta}$ and $\nicefrac{\partial\Sigma}{\partial\theta}$ to
compute the gradient of the negative log-likelihood w.r.t. the $g$-space \acro{GP} 
hyperparamters via the chain rule.

The relevant partial derivatives for the log transform are trivial to compute:
\begin{align}
	\frac{\partial\E[x_i]}{\partial\mu_i}&=\exp\left(\mu_i+\frac{\Sigma_{ii}}{2}\right) \\
	\frac{\partial\E[x_i]}{\partial\Sigma_{ii}}&=\frac{1}{2}\exp\left(\mu_i+\frac{\Sigma_{ii}}{2}\right) \\
	\frac{\partial\E[x_i^2]}{\partial\mu_i}&=2\exp\left(2\mu_i+2\Sigma_{ii}\right) \\
	\frac{\partial\E[x_i^2]}{\partial\Sigma_{ii}}&=2\exp\left(2\mu_i+2\Sigma_{ii}\right) \\
	\frac{\partial\E[x_ix_j]}{\partial\mu_i}&=\exp\left(\mu_i+\mu_j+\frac{1}{2}(\Sigma_{ii}+\Sigma_{jj})+\Sigma_{ij}\right) \\
	\frac{\partial\E[x_ix_j]}{\partial\Sigma_{ii}}&=\frac{1}{2}\exp\left(\mu_i+\mu_j+\frac{1}{2}(\Sigma_{ii}+\Sigma_{jj})+\Sigma_{ij}\right) \\
	\frac{\partial\E[x_ix_j]}{\partial\Sigma_{ij}}&=\exp\left(\mu_i+\mu_j+\frac{1}{2}(\Sigma_{ii}+\Sigma_{jj})+\Sigma_{ij}\right)
	\label{eq0}
\end{align}

\subsection{Probit transform moments}
To derive the first raw moment associated with the probit transform, we take an
approach similar to the one found in section 3.9 of \citet{rasmussen06}: let
$\vec{y}=\{y_1,\ldots,y_n\}$ be a multivariate Gaussian random variable with
mean vector $\vec{\mu}$ and covariance matrix $\Sigma$ and let $\vec{x}=\Phi(\vec{y})$.
The first raw moment of $x_i$ is
\begin{align}
	\E[x_i]&=\int_{-\infty}^{\infty}\Phi(w)\phi(w,\mu_i,\Sigma_{ii})\intd{w} \\
		     &=\frac{1}{2\pi\sqrt{\Sigma_{ii}}}\int_{-\infty}^{\infty}\int_{-\infty}^{w}\exp\left(-\frac{z^2}{2}-\frac{(w-\mu_i)^2}{2\Sigma_{ii}}\right)\intd{z}\intd{w}.
	\label{eq1}
\end{align}
We make the following substitutions: $a=w-\mu_i$ and $b=z-a$. Plugging these
substitutions in and then switching the order of the integrals gives
\begin{equation}
	\E[x_i]=\frac{1}{2\pi\sqrt{\Sigma_{ii}}}\int_{-\infty}^{\mu_i}\int_{-\infty}^{\infty}\exp\left(-\frac{(a+b)^2}{2}-\frac{a^2}{2\Sigma_{ii}}\right)\intd{a}\intd{b}.
	\label{eq2}
\end{equation}
Observe that the quantity inside the exponent of \eqref{eq2} can be written
using matrix notation as
\begin{align}
  (a+b)^2&+\frac{a^2}{\Sigma_{ii}}=\begin{bmatrix}a & b\end{bmatrix}\begin{bmatrix}1+\frac{1}{\Sigma_{ii}} & 1 \\ 1 & 1\end{bmatrix}\begin{bmatrix}a \\ b\end{bmatrix} \\
										&=\begin{bmatrix}a & b\end{bmatrix}\begin{bmatrix}\Sigma_{ii} & -\Sigma_{ii} \\ -\Sigma_{ii} & 1+\Sigma_{ii}\end{bmatrix}\inv\begin{bmatrix}a \\ b\end{bmatrix},
	\label{eq3}
\end{align}
thus revealing the integrand of \eqref{eq2} to be (proportional to) a bivariate
Gaussian \acro{PDF}. The innermost integral of \eqref{eq2} is therefore equivalent
to marginalizing out one of the variables in this bivariate distribution, up to a
normalizing constant which can be pulled from the constants in front of the integral.
Continuing the derivation in this way gives
\begin{align}
	\E[x_i]&=\frac{1}{\sqrt{2\pi(1+\Sigma_{ii})}}\int_{-\infty}^{\mu_i}\exp\left(-\frac{b^2}{2(1+\Sigma_{ii})}\right)\intd{b} \\
			   &=\Phi(\mu_i,0,1+\Sigma_{ii})=\Phi\left(\frac{\mu_i}{\sqrt{1+\Sigma_{ii}}}\right).
	\label{eq4}
\end{align}
To derive the second raw moments associated with the probit transform, we begin
with an approach similar to the one above. We start with the product moment
$\E[x_ix_j]$ (for notational simplicity, let $\Sigma_{(i,j)}=\begin{bsmallmatrix}
\Sigma_{ii} & \Sigma_{ij} \\ \Sigma_{ji} & \Sigma_{jj}\end{bsmallmatrix}$):
\begin{align}
	\E&[x_ix_j]=\int_{-\infty}^{\infty}\int_{-\infty}^{\infty}\Phi(w_1)\Phi(w_2) \nonumber \\
	  &\phi\left(\begin{bmatrix}w_1 \\ w_2\end{bmatrix},\begin{bmatrix}\mu_i \\ \mu_j\end{bmatrix},\Sigma_{(i,j)}\right)\intd{w_1}\intd{w_2} \\
		&=\frac{1}{4\pi^2|\Sigma_{(i,j)}|^{\nicefrac{1}{2}}}\int_{-\infty}^{\infty}\int_{-\infty}^{\infty}\int_{-\infty}^{w_1}\int_{-\infty}^{w_2} \nonumber \\
		&\exp\left(-\frac{1}{2}\left(\begin{bmatrix}w_1-\mu_i \\ w_2-\mu_j\end{bmatrix}\trans\Sigma_{(i,j)}\inv\begin{bmatrix}w_1-\mu_1 \\ w_2-\mu_2\end{bmatrix}\right)\right) \nonumber \\
		&\exp\left(-\frac{1}{2}\left(z_1^2+z_2^2\right)\right)\intd{z_2}\intd{z_1}\intd{w_2}\intd{w_1}.
	\label{eq5}
\end{align}
Next, we make the following substitutions: $a_1=w_1-\mu_i$, $a_2=w_2-\mu_j$, $b_1=z_1-a_1$ and $b_2=z_2-a_2$:
\begin{align}
	\E&[x_ix_j]=\frac{1}{4\pi^2|\Sigma_{(i,j)}|^{\nicefrac{1}{2}}} \nonumber \\
	  &\int_{-\infty}^{\mu_i}\int_{-\infty}^{\mu_j}\int_{-\infty}^{\infty}\int_{-\infty}^{\infty}\exp\left(-\frac{1}{2}\left(\begin{bmatrix}a_1 \\ a_2\end{bmatrix}\trans\Sigma_{(i,j)}\inv\begin{bmatrix}a_1 \\ a_2\end{bmatrix}\right)\right) \nonumber \\
		&\exp\left(-\frac{1}{2}\left((a_1+b_1)^2+(a_2+b_2)^2\right)\right)\intd{a_2}\intd{a_1}\intd{b_2}\intd{b_1}.
	\label{eq6}
\end{align}
We can again express the exponent in \eqref{eq6} using matrix notation as follows
\begin{align}
	 &\begin{bmatrix}a_1 \\ a_2\end{bmatrix}\trans\Sigma_{(i,j)}\inv\begin{bmatrix}a_1 \\ a_2\end{bmatrix}+(a_1+b_1)^2+(a_2+b_2)^2 \nonumber \\
	=&\begin{bmatrix}a_1 \\ a_2 \\ b_1 \\ b_2\end{bmatrix}\trans
		\begin{bmatrix} \Sigma_{(i,j)}\inv & I_2 \\ I_2 & I_2 \end{bmatrix}
		\begin{bmatrix}a_1 \\ a_2 \\ b_1 \\ b_2\end{bmatrix},
	\label{eq7}
\end{align}
where $I_2$ is the 2-by-2 identity matrix. In this form, we can recognize the integrand of \eqref{eq6} to be
proportional to a multivariate Gaussian \acro{PDF}. Pulling constants from outside the integral gives
\begin{align}
 \E&[x_ix_j]=\int_{-\infty}^{\mu_i}\int_{-\infty}^{\mu_j}\int_{-\infty}^{\infty}\int_{-\infty}^{\infty} \nonumber \\
   &\phi\left(\begin{bmatrix}a_1 \\ a_2 \\ b_1 \\ b_2\end{bmatrix},\begin{bmatrix}0 \\ 0 \\ 0 \\ 0\end{bmatrix},\begin{bmatrix}\Sigma_{(i,j)} & -\Sigma_{(i,j)} \\ -\Sigma_{(i,j)} & \Sigma_{(i,j)}+I_2 \end{bmatrix}\right)\intd{a_2}\intd{a_1}\intd{b_2}\intd{b_1}.
 \label{eq8}
\end{align}
Thus, the two innermost integrals correspond to marginalizing out the variable $a_1$ and $a_2$ from this
multivariate Gaussian and so we arrive at the final result:
\begin{align}
 \E[x_ix_j]&=\int_{-\infty}^{\mu_i}\int_{-\infty}^{\mu_j}\phi\left(\begin{bmatrix}b_1 \\ b_2\end{bmatrix},\begin{bmatrix}0 \\ 0\end{bmatrix},\Sigma_{(i,j)}+I_2\right)\intd{b_2}\intd{b_1} \nonumber \\
   				 &=\Phi\left(\begin{bmatrix}\mu_i \\ \mu_j\end{bmatrix},\Sigma_{(i,j)}+I_2\right).
 \label{eq9}
\end{align}
Using the same derivation as detailed above, we can show that
\begin{equation}
 \E[x_i^2]=\Phi\left(\begin{bmatrix}\mu_i \\ \mu_i\end{bmatrix},\begin{bmatrix}\Sigma_{ii}+1 & \Sigma_{ii} \\ \Sigma_{ii} & \Sigma_{ii}+1\end{bmatrix}\right).
 \label{eq10}
\end{equation}
Again, we present the relevant partial derivatives, starting with the partial derivatives of the first moment:
\begin{align}
  \frac{\partial\E[x_i]}{\partial\mu_i}&=\phi\left(\frac{\mu_i}{\sqrt{1+\Sigma_{ii}}}\right)\frac{1}{\sqrt{1+\Sigma_{ii}}} \\
  \frac{\partial\E[x_i]}{\partial\Sigma_{ii}}&=\phi\left(\frac{\mu_i}{\sqrt{1+\Sigma_{ii}}}\right)\frac{-\mu_i}{2(1+\Sigma_{ii})^{\nicefrac{3}{2}}},
  \label{eq11}
\end{align}
which follow from the fundamental theorm of calculus and the chain rule. The derivative of the second raw
moments w.r.t. $\mu_i$ can also be computed in a similar fashion:
\begin{align}
  \frac{\partial\E[x_i^2]}{\partial\mu_i}&=\frac{\partial}{\partial\mu_i}\Phi\left(\begin{bmatrix}\mu_i \\ \mu_i\end{bmatrix},\begin{bmatrix}\Sigma_{ii}+1 & \Sigma_{ii} \\ \Sigma_{ii} & \Sigma_{ii}+1\end{bmatrix}\right) \nonumber \\
                                             &=\int_{-\infty}^{\mu_i}\phi\left(\begin{bmatrix}b_1 \\ \mu_i\end{bmatrix},\begin{bmatrix}0 \\ 0\end{bmatrix},\Sigma_{(i,j)}+I_2\right)\intd{b_1} \nonumber \\
                                             &+\int_{-\infty}^{\mu_i}\phi\left(\begin{bmatrix}\mu_i \\ b_2\end{bmatrix},\begin{bmatrix}0 \\ 0\end{bmatrix},\Sigma_{(i,j)}+I_2\right)\intd{b_2} \nonumber \\
                                             &=2\int_{-\infty}^{\mu_i}\phi\left(\begin{bmatrix}\mu_i \\ b\end{bmatrix},\begin{bmatrix}0 \\ 0\end{bmatrix},\Sigma_{(i,j)}+I_2\right)\intd{b} \nonumber \\
                                             &=2\bigg(\phi\left(\frac{\mu_i}{\sqrt{1+\Sigma_{ii}}}\right) \nonumber \\
                                             &\quad\phantom{--}\Phi\left(\mu_i,\frac{\Sigma_{ii}}{\Sigma_{ii}+1}\mu_i,\frac{\Sigma_{ii}}{\Sigma_{ii}+1}+1\right)\bigg)
  \label{eq12}
\end{align}
where the last line can be arrived at by pulling the term $\phi\left(\nicefrac{\mu_i}{\sqrt{1+\Sigma_{ii}}}\right)$
out of the integral and then completing the square. Following a similar derivation, the partial derivative of the
product moment w.r.t. $\mu_i$ is
\begin{align}
  \frac{\partial\E[x_ix_j]}{\partial\mu_i}&=\phi\left(\frac{\mu_i}{\sqrt{1+\Sigma_{ii}}}\right) \nonumber \\
                                              &\phantom{iiii}\Phi\left(\mu_j,\frac{\Sigma_{ij}}{\Sigma_{ii}+1}\mu_i,\Sigma_{jj}+1-\frac{\Sigma_{ij}^2}{\Sigma_{ii}+1}\right).
  \label{eq13}
\end{align}
Next, the partial derivative of the second moment w.r.t. the covariance can be computed as follows:
\begin{align}
  \frac{\partial\E[x_ix_j]}{\partial\Sigma_{ij}}=-\frac{\Sigma_{ij}}{|\Sigma_{(i,j)}|}\bigg(&\Phi\left(\begin{bmatrix}\mu_i \\ \mu_j\end{bmatrix},\Sigma_{(i,j)}+I_2\right) \nonumber \\
                                               +\int_{-\infty}^{\mu_i}\int_{-\infty}^{\mu_j}&\phi\left(\begin{bmatrix}\mu_i \\ \mu_j\end{bmatrix},\Sigma_{(i,j)}+I_2\right) \nonumber \\
                                             &\hspace{-1cm}\bigg(\begin{bmatrix}b_1 \\ b_2\end{bmatrix}\trans\Sigma_{(i,j)}\inv\begin{bmatrix}b_1 \\ b_2\end{bmatrix}-\frac{b_1b_2}{\Sigma_{ij}}\bigg)\intd{b_2}\intd{b_1}\bigg).
  \label{eq14}
\end{align}
The second term can be decomposed into a weighted sum of the second raw moments of a truncated bivariate Gaussian.
These moments can be expressed in terms of the univariate Gaussian \acro{PDF} and \acro{CDF} \citep{rosenbaum61}:
\begin{align}
  \int_{-\infty}^{\mu_i}\int_{-\infty}^{\mu_j}&b_1^2\phi\left(\begin{bmatrix}b_1 \\ b_2\end{bmatrix},\Sigma_{(i,j)}+I_2\right)\intd{b_2}\intd{b_1} \nonumber \\
                                              &=(\Sigma_{ii}+1)\bigg(\Phi\left(\begin{bmatrix}\mu_i \\ \mu_j\end{bmatrix},\Sigma_{(i,j)}+I_2\right) \nonumber \\
                                              &-t_1(\mu_i^*,\mu_j^*,\rho)-\rho^2t_2(\mu_i^*,\mu_j^*,\rho) \nonumber \\
                                              &+\rho\sqrt{1-\rho^2}t_3(\mu_i^*,\mu_j^*,\rho)\bigg)
  \label{eq15}
\end{align}
where
\begin{align}
  \rho&=\frac{\Sigma_{ij}^2}{(\Sigma_{ii}+1)(\Sigma_{jj}+1)} \\
  \mu_i^*&=\frac{\mu_i}{\Sigma_{ii}+1} \\
  \mu_j^*&=\frac{\mu_j}{\Sigma_{jj}+1} \\
  t_1(\mu_i^*,\mu_j^*,\rho)&=\mu_i^*\phi\left(\mu_i^*\right)\Phi\left(\frac{\mu_j^*-\rho\mu_i^*}{\sqrt{1-\rho^2}}\right) \\
  t_2(\mu_i^*,\mu_j^*,\rho)&=\mu_j^*\phi\left(\mu_j^*\right)\Phi\left(\frac{\mu_i^*-\rho\mu_j^*}{\sqrt{1-\rho^2}}\right) \\
  t_3(\mu_i^*,\mu_j^*,\rho)&=\frac{1}{\sqrt{2\pi}}\phi\left(\sqrt{\frac{\mu_i^{*2}-2\rho\mu_i^*mu_j^*+\mu_j^{*2}}{1-\rho^2}}\right).
  \label{eq16}
\end{align}
Similarly, the other moments of the truncated bivariate Gaussian are
\begin{align}
  \int_{-\infty}^{\mu_i}\int_{-\infty}^{\mu_j}&b_2^2\phi\left(\begin{bmatrix}b_1 \\ b_2\end{bmatrix},\Sigma_{(i,j)}+I_2\right)\intd{b_2}\intd{b_1} \nonumber \\
                                              &=(\Sigma_{jj}+1)\bigg(\Phi\left(\begin{bmatrix}\mu_i \\ \mu_j\end{bmatrix},\Sigma_{(i,j)}+I_2\right) \nonumber \\
                                              &-\rho^2t_1(\mu_i^*,\mu_j^*,\rho)-t_2(\mu_i^*,\mu_j^*,\rho) \nonumber \\
                                              &+\rho\sqrt{1-\rho^2}t_3(\mu_i^*,\mu_j^*,\rho)\bigg)
  \label{eq17}
\end{align}
and
\begin{align}
  \int_{-\infty}^{\mu_i}\int_{-\infty}^{\mu_j}&b_1b_2\phi\left(\begin{bmatrix}b_1 \\ b_2\end{bmatrix},\Sigma_{(i,j)}+I_2\right)\intd{b_2}\intd{b_1} \nonumber \\
                                              &=\rho\sqrt{(\Sigma_{ii}+1)(\Sigma_{jj}+1)} \nonumber \\
                                              &\bigg(\Phi\left(\begin{bmatrix}\mu_i \\ \mu_j\end{bmatrix},\Sigma_{(i,j)}+I_2\right)-t_1(\mu_i^*,\mu_j^*,\rho) \nonumber \\
                                              &-t_2(\mu_i^*,\mu_j^*,\rho)+\frac{\sqrt{1-\rho^2}}{\rho}t_3(\mu_i^*,\mu_j^*,\rho)\bigg).
  \label{eq18}
\end{align}
We can therefore substitute \eqref{eq15}, \eqref{eq17} and \eqref{eq18} into \eqref{eq14} to come up with a closed form for
this partial derivative. Lastly, the partial derivatives of the second raw moments w.r.t. $\Sigma_{ii}$ are
\begin{align}
  \frac{\partial\E[x_ix_j]}{\partial\Sigma_{ii}}=-\frac{\Sigma_{jj}+1}{2|\Sigma_{(i,j)}|}\bigg(&\Phi\left(\begin{bmatrix}\mu_i \\ \mu_j\end{bmatrix},\Sigma_{(i,j)}+I_2\right) \nonumber \\
                                                  +\int_{-\infty}^{\mu_i}\int_{-\infty}^{\mu_j}&\phi\left(\begin{bmatrix}\mu_i \\ \mu_j\end{bmatrix},\Sigma_{(i,j)}+I_2\right) \nonumber \\
                                             &\hspace{-1.5cm}\bigg(\begin{bmatrix}b_1 \\ b_2\end{bmatrix}\trans\Sigma_{(i,j)}\inv\begin{bmatrix}b_1 \\ b_2\end{bmatrix}+\frac{b_2^2}{\Sigma_{jj}+1}\bigg)\intd{b_2}\intd{b_1}\bigg)
  \label{eq19}
\end{align}
and
\begin{align}
  \frac{\partial\E[x_i^2]}{\partial\Sigma_{ii}}=-\frac{1}{2\Sigma_{ii}+1}\bigg(&\Phi\left(\begin{bmatrix}\mu_i \\ \mu_i\end{bmatrix},\begin{bmatrix}\Sigma_{ii}+1 & \Sigma_{ii} \\ \Sigma_{ii} & 1+\Sigma_{ii}\end{bmatrix}\right) \nonumber \\
                                  -\int_{-\infty}^{\mu_i}\int_{-\infty}^{\mu_i}&\phi\left(\begin{bmatrix}\mu_i \\ \mu_i\end{bmatrix},\begin{bmatrix}\Sigma_{ii}+1 & \Sigma_{ii} \\ \Sigma_{ii} & 1+\Sigma_{ii}\end{bmatrix}\right) \nonumber \\
                                             &\hspace{-2.4cm}\bigg(\begin{bmatrix}b_1 \\ b_2\end{bmatrix}\trans\begin{bmatrix}\Sigma_{ii} & \Sigma_{ii}+1\\\Sigma_{ii}+1 & \Sigma_{ii}\end{bmatrix}\inv\begin{bmatrix}b_1 \\ b_2\end{bmatrix}\bigg)\intd{b_2}\intd{b_1}\bigg).
  \label{eq20}
\end{align}
We can again substitute \eqref{eq15}, \eqref{eq17} and \eqref{eq18} into \eqref{eq19} and \eqref{eq20}
to come up with closed forms for these partial derivatives.

\newpage
\textrm{ }
\newpage

\bibliography{main}

\end{document}